\documentclass[conference]{IEEEtran} %
\IEEEoverridecommandlockouts
\usepackage{cite}
\usepackage{amsmath,amssymb,amsfonts}
\usepackage{algorithmic}
\usepackage{graphicx}
\usepackage{textcomp}
\usepackage{xcolor}
\def\BibTeX{{\rm B\kern-.05em{\sc i\kern-.025em b}\kern-.08em
    T\kern-.1667em\lower.7ex\hbox{E}\kern-.125emX}}
\begin{document}

\title{Multi-task Gaze Estimation Via Unidirectional Convolution\\
%{\footnotesize \textsuperscript{*}Note: Sub-titles are not captured for https://ieeexplore.ieee.org  and
}

\author{\IEEEauthorblockN{Zhang Cheng}
\IEEEauthorblockA{\textit{College of Computer and Information Science} \\
\textit{Chongqing Normal University}\\
Chongqing, China\\
2022210516040@stu.cqnu.edu.cn}
\and
%\IEEEauthorblockN{Yanxia Wang\textsuperscript{*}}
\IEEEauthorblockN{Yanxia Wang* \thanks{ * is corresponding author.}}
\IEEEauthorblockA{\textit{College of Computer and Information Science} \\
\textit{Chongqing Normal University}\\
Chongqing, China\\
wangyanxia@cqnu.edu.cn}
}

\maketitle

\begin{abstract}
Using lightweight models as backbone networks in gaze estimation tasks often results in significant performance degradation. The main reason is that the number of feature channels in lightweight networks is usually small, which makes the model expression ability limited. In order to improve the performance of lightweight models in gaze estimation tasks, a network model named Multitask-Gaze is proposed. The main components of Multitask-Gaze include Unidirectional Convolution (UC), Spatial and Channel Attention (SCA), Global Convolution Module (GCM), and Multi-task Regression Module(MRM). UC not only significantly reduces the number of parameters and FLOPs, but also extends the receptive field and improves the long-distance modeling capability of the model, thereby improving the model performance. SCA highlights gaze-related features and suppresses gaze-irrelevant features. The GCM replaces the pooling layer and avoids the performance degradation due to information loss. MRM improves the accuracy of individual tasks and strengthens the connections between tasks for overall performance improvement. The experimental results show that compared with the State-of-the-art method SUGE, the performance of Multitask-Gaze on MPIIFaceGaze and Gaze360 datasets is improved by 1.71\% and 2.75\%, respectively, while the number of parameters and FLOPs are significantly reduced by 75.5\% and 86.88\%.
\end{abstract}


\section{Introduction}
Gaze estimation reveals human intention information and points of interest, and is widely used in various fields such as autonomous driving \cite{b1}, human-computer interaction\cite{b2}, psychological research \cite{b3}, and medical diagnosis \cite{b4}\cite{b5}. The gaze estimation methods can be divided into two types: model-based \cite{b6}\cite{b7} and appearance-based \cite{b8}\cite{b9}. Model-based methods rely on eye structure models and can achieve high accuracy, but this method requires specialized equipment and has limited usage scenarios. The appearance-based method is less restricted by the environment and has a simple device, and deep learning techniques can be used to predict gaze direction from direct facial or eye images. In recent years, appearance-based methods have gradually become a research hotspot due to their convenience and wide application scenarios.

However, current appearance-based methods mostly improve model performance by increasing network depth, while performance degrades when performing gaze estimation tasks using lightweight networks. In order to improve the above problems, a variety of lightweight networks such as FR-Net\cite{b10} and GazeNAS-ETH\cite{b11} have been proposed. However, FR-Net uses higher resolution images as input data, which increases the computational burden and reduces the real-time performance of the model. GazeNAS-ETH requires parameter search of the model on a large dataset to achieve advanced performance. Lightweight models have a wider range of application scenarios, but are also more challenging because it is difficult to balance the number of model parameters and accuracy. Therefore, lightweight model still has important research significance.

An important reason for the performance degradation of lightweight models is that the number of feature channels is small, which leads to the lack of fitting ability. In addition, lightweight models usually use deep convolution \cite{b12} instead of standard convolution, because the lack of inter-channel information fusion in deep convolution reduces the model's expressiveness. One possible solution is to fuse attention mechanism in the network, make it pay more attention to gaze-related information.

In this paper, we propose a lightweight multi-task gaze estimation model for the above problem. This model not only further reduces the parameters of the existing lightweight model, but also better balances the number of parameters and performance of the model.

The main contributions of this paper are as follows: 
\begin{itemize}
\item A Unidirectional Convolution (UC) is proposed. UC replaces deep convolution, which not only significantly reduces the number of parameters and FLOPs, but also extends the receptive field of the model to improve the long-distance modeling capability and hence the model performance.

\begin{figure*}[htbp]
\centerline{\includegraphics[width=0.95\linewidth]{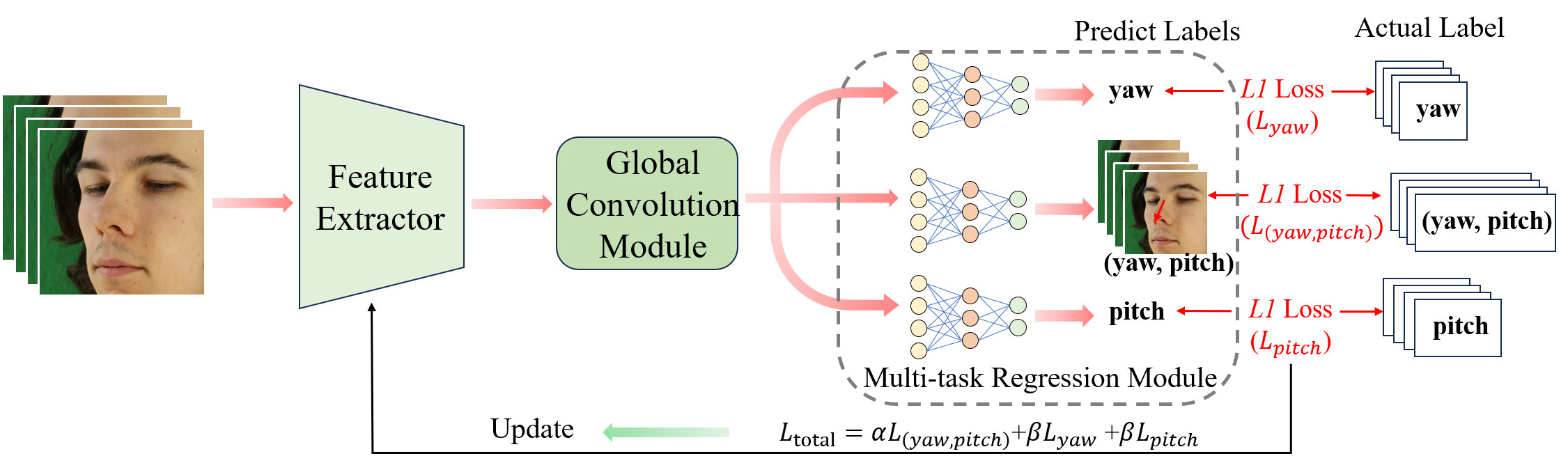}}
\caption{Structure of Multitask-Gaze}
\label{Multitask}
\end{figure*}

\item A new Spatial and Channel Attention (SCA) is proposed. spatial attention realizes global information perception at the spatial level, making full use of context information and increasing the weight of important information in space. channel attention realizes information interaction between channels. 
\item A Global Convolutional Module (GCM) based on UC is proposed. GCM replaces the pooling layer to further integrate global information and avoid performance degradation caused by information loss. 
\item A Multi-task Regression Module (MRM) is proposed. The model improves the performance of individual tasks and strengthens the correlation between tasks, thereby improving the overall performance.
\end{itemize}

\section{Methods}
This section details the proposed Multitask-Gaze network model, which consists of three components: Feature Extractor, Global Convolution Module, and Multi-task Regression Module. The structure is shown in Fig. 1.

\subsection{Feature Extractor}\label{AA}
The feature extractor is improved based on MobileNetV3\cite{b13}, and the main difference from the original MobileNetV3 is the use of UC to improve the bneck block, while adding SCA after the 3rd, 6th, and 9th bneck blocks, respectively.
\begin{figure}[htbp]
\centerline{\includegraphics[width=0.8\linewidth]{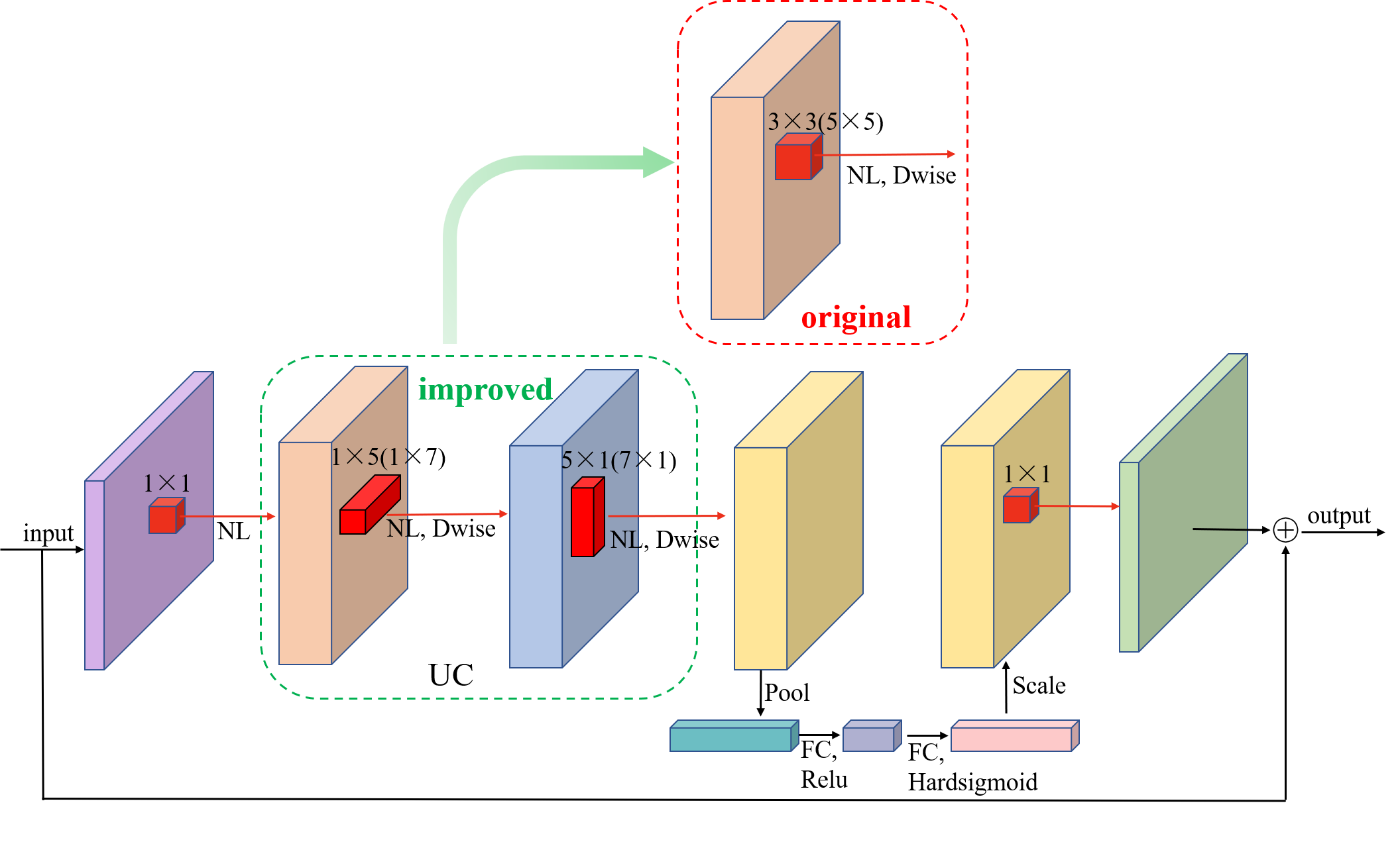}}
\caption{Structure of bneck}
\label{bneck}
\end{figure}

\textbf{Unidirectional Convolution(UC).} 
Bneck is the basic building block of a network, and its structure is shown in Fig. 2. In this paper, we improve the bneck module with UC to make it more lightweight. In order to expand the receptive field of the model and fully integrate contextual information to improve performance, this paper replaces the 3×3 or 5×5 convolutions in the original bneck structure with UC of [1×5;5×1] or [1×7; 7×1], respectively. Assuming the size of the output feature map is $H_{out}*W_{out}$, the convolution kernel is $K_{H}*K_{W}$, the input channel is $C_{in}$, and the output channel is $C_{out}$, the parameter and computational complexity of the improved part in bneck are as follows: 
\begin{equation}
\label{eq1}
\begin{split}
  &Para_{original}=\ K_{H}*K_{W}*C_{in}*C_{out} \\
 &Para_{UC}=\ 1*K_{W}*C_{in}*C_{out}+K_{H}*1*C_{in}*C_{out} \\
 &FLOPs_{original}=\ (K_{H}*K_{W}*C_{in}*C_{out})*H_{out}*W_{out} \\
 &FLOPs_{UC} =(1*K_{W}*C_{in}*C_{out})*H_{out}*W_{out}   
           \\&\quad\quad\quad\quad+(K_{H}*1*C_{in}*C_{out})*H_{out}*W_{out}
\end{split}
\end{equation}
When both the input and output channels are 40, the output feature map size is 56×56, and the original convolution kernel is 5×5, UC reduces the number of parameters and FLOPs by at least 44\%.

\textbf{Spatial and Channel Attention(SCA).} 
In lightweight networks, standard convolutions are usually replaced with deep convolutions to achieve lightweighting, but there is no information exchange between deep convolution channels, resulting in performance degradation. In order to compensate for the shortcomings of deep convolution, SCA is proposed in this paper, whose structure is shown in Fig. 3.  SCA is divided into two parts: Spatial Attention and Channel Attention. Spatial Attention not only highlights important features at the spatial level, but also achieves global spatial information exchange through the shift window mechanism \cite{b14}, enhancing the model's long-distance modeling capability. The Channel Attention further integrates global information through average pooling, extracts the most significant features through max-pooling, and then uses MLP to complete information exchange between channels.
\begin{figure}[htbp]
\centerline{\includegraphics[width=0.85\linewidth]{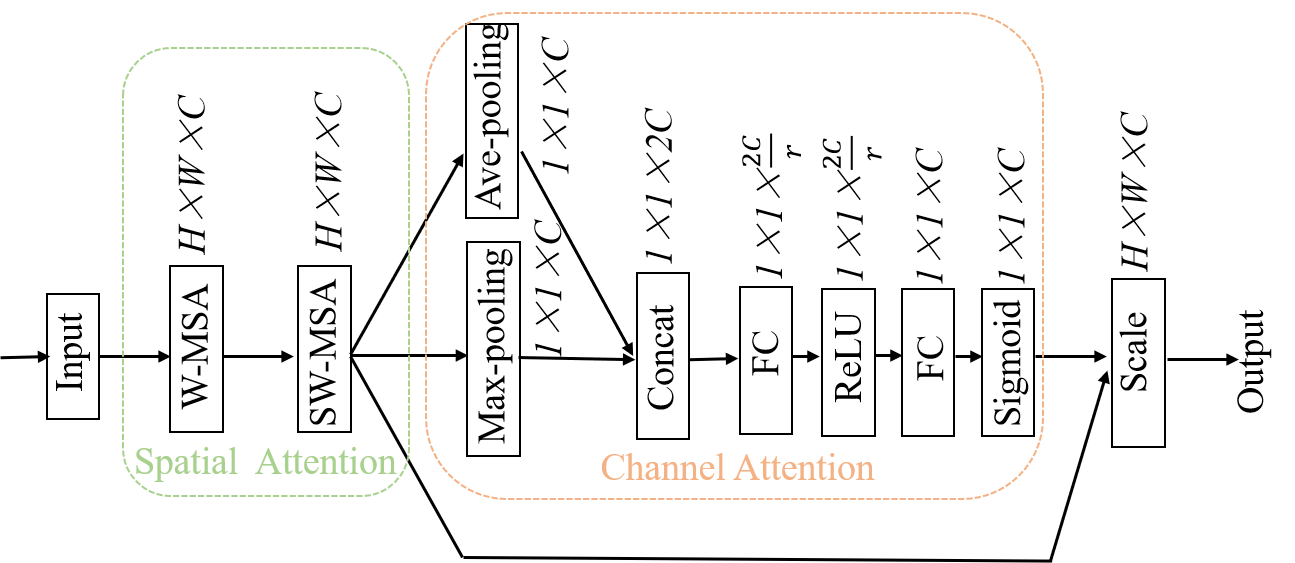}}
\caption{Structure of SCA}
\label{SCA}
\end{figure}

\subsection{Global Convolution Module }
In CNN, when classifying or regressing various tasks, it is usually necessary to first perform global information fusion on the feature maps through global average pooling. However, global average pooling will to some extent ignore local details and differences in the feature map, and if important information is lost, it will lead to performance degradation. On the other hand, if there are outliers in the feature map, they may have a significant impact on the results of global average pooling, thereby reducing the robustness and accuracy of the model. In terms of the above issues, this paper proposes a Global Convolution Module (GCM) based on UC, which integrates global information without causing information loss. Its structure is shown in Fig. 4.
\begin{figure}[htbp]
\centerline{\includegraphics[width=0.8\linewidth]{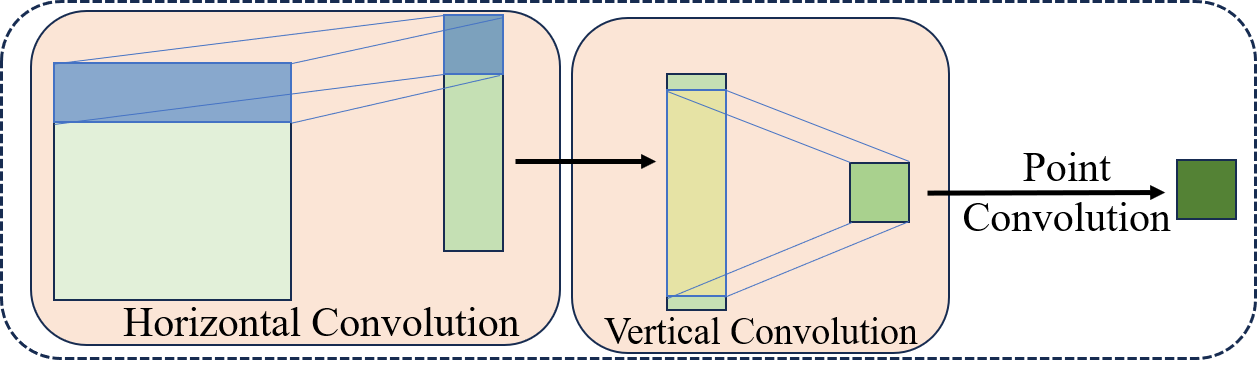}}
\caption{Structure of GCM}
\label{GCM}
\end{figure}

\begin{figure}[htbp]
\centerline{\includegraphics[width=0.8\linewidth]{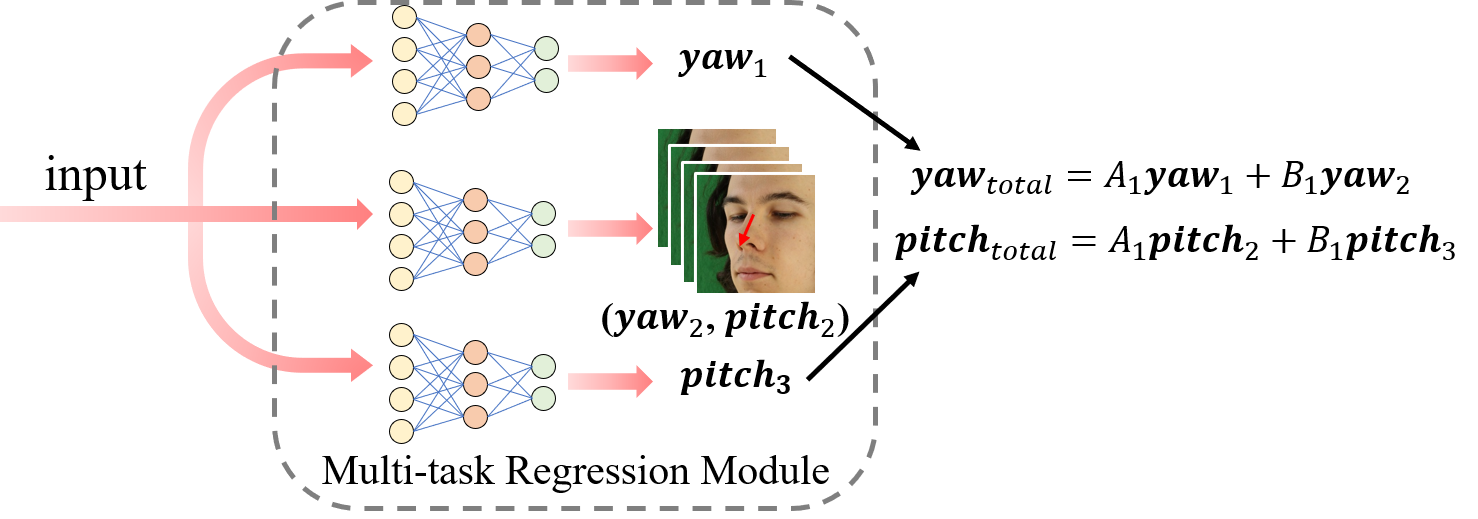}}
\caption{Structure of MRM}
\label{MRM}
\end{figure}

\subsection{Multi-task Regression Module}
The function of the Multi-task Regression Module (MRM) is to predict yaw, (yaw, pitch) and pitch respectively, by using the feature $f\in\mathbb{R}^{1\times1\times480}$ extracted by GCM. Its structure is shown in Fig. 5. The MRM consists of three MLP modules, with two MLP modules predicting yaw and pitch, respectively, to improve the accuracy of a single Angle. However, human gaze change is usually a process of continuous change, there is a certain correlation between yaw and pitch, and individual prediction may not be able to make full use of the intrinsic connection and interdependence between yaw and pitch, thus limiting the comprehensive performance. To solve this problem, this paper adds another MLP module to predict yaw and pitch at the same time, and finally assigns different weight parameters to yaw and pitch through hyperparameter setting. The calculation formula is as follows:
\begin{equation}
\label{eq2}
\begin{split}
  &(yaw_{1}, pitch_{1})=\ (MLP_{1}(f), MLP_{3}(f)) \\
  &(yaw_{2}, pitch_{2})=\ MLP_{2}(f) \\
  &yaw_{total}=\ A_{1}yaw_{1} + B_{1}yaw_{2} \\
  &pitch_{total}=\ A_{1}pitch_{1} + B_{1}pitch_{2} 
\end{split}
\end{equation}
Where $A_{1}$= $B_{1}$=0.5, $f$ represents the feature input to MRM.

\section{Experimental}

\subsection{Implementation Details}
In this paper, four datasets of ETH-XGaze\cite{b15}, Gaze360\cite{b16}(G), MPIIFaceGaze\cite{b17}(MPII), and RT-Gene\cite{b18}(RT) were used to verify the performance of Multitask-Gaze model. ETH-Xgaze is used as the pre-training dataset. In order to effectively avoid overfitting of low-level features and loss of high-level semantics, a linear decay random dropout rate\cite{b19} is added to the three SCA modules during training. The parameter number, FLOPs, and Angle error(the smaller angular error indicates the better model performance) are used as evaluation metrics in this paper, and the Angle error is calculated as follows.
\begin{equation}
\label{eq3}
L_{angular}=\arccos(\frac{g\bullet\hat{g}}{\parallel g \parallel\parallel\hat{g}\parallel\ \ })
\end{equation}
Among them, $g\in\mathbb{R}^3$ represents the actual gaze direction, $\hat{g}\in\mathbb{R}^3$ represents the predicted gaze direction.

The loss function adopts MAE Loss (Mean Absolute Error), and the calculation formula is as follows:
\begin{equation}
\label{eq4}
\begin{split}
  &(L_{\text {yaw }},L_{\text {pitch }})=(\frac{1}{N} \sum_{i=1}^{N}\left|g_{\text {yaw }}^{i}-\hat{g}_{\text {yaw }}^{i}\right|,\frac{1}{N} \sum_{i=1}^{N}\left|g_{\text {pitch }}^{i}-\hat{g}_{\text {pitch }}^{i}\right|) \\
  &L_{\text {(yaw,pitch) }}=\frac{1}{N} \sum_{i=1}^{N}\left|g_{(\text {(yaw, pitch)})}^{i}-\hat{g}_{(\text {yaw }, \text { pitch })}^{i}\right|\\
  &L_{\text {total}}=A_{2}L_{\text {(yaw,pitch)}}+B_{2}L_{\text {yaw }}+B_{2}L_{\text {pitch }}
\end{split}
\end{equation}
Where $A_{2}$=0.5,$B_{2}$=0.5, $N$ represents the number of samples.

\begin{table}[]
\caption{Comparison with State-of-the-art methods.}
\scalebox{0.8}{
\begin{tabular}{lllllll}
\hline
Methods              & Params(M) & FLOPs(G) & G\cite{b16} & MPII\cite{b17} & RT\cite{b18} & \multicolumn{1}{l}{Year} \\ \hline
Dilated-Net\cite{b20}  & 3.92      & 3.15     & 13.73°    & 4.42°        & 8.38°      & 2018                      \\
Gaze360\cite{b16}      & 11.90     & 7.29     & 11.04°    & 4.06°        & 7.06°      & 2019                      \\
GazeTR\cite{b21}       & 11.42     & 1.83     & 10.62°    & 4.00°        & 6.55°      & 2022                      \\
CADSE\cite{b22}        & 74.80     & 19.75    & 10.70°    & 4.04°        & 7.00°      & 2022                      \\
L2CS-Net\cite{b23}     & 23.52     & 16.53    & 10.41°    & 3.92°        & N/A        & 2023                      \\
GazeCaps\cite{b24}     & 11.70     & 1.82     & 10.04°    & 4.06°        & 6.92°      & 2023                      \\
Gaze-Swin\cite{b25}    & 32.28     & 5.17     & 10.14°    & N/A          & 6.38°      & 2024                      \\
SUGE\cite{b26}        & 11.42     & 1.83     & 10.51°    & 4.01°        & N/A        & 2024                      \\ \hline
MobileNetV3\cite{b13}* & 4.205     & 0.233    & 10.46     & 4.05         & 6.37       &                           \\
Multitask-Gaze(Our)  & 2.822     & 0.243    & 10.33°    & 3.90°        & 6.35°      &                           \\ \hline
\multicolumn{7}{l}{* indicates performance applied to gaze estimation tasks, not original paper data}
\end{tabular}}
\end{table}

\subsection{Comparison with State-of-the-art methods}
This section compares Multitask-Gaze with existing state-of-the-art models in terms of angular error, parameters, and FLOPs, and the results are shown in Table 1.As can be seen from Table 1, although the Angle error of Gaze360 dataset by Gazet-Swin \cite{b25} is 0.19°(1.84\%) lower than that of Multitask-Gaze, the parameter number and FLOPs of Multitask-Gaze are reduced by 91.2\% and 95.36\% respectively. Compared with the original MobileNetV3, Multitask-Gaze has a 0.13°(1.24\%), 0.15(3.7\%) and 0.02(0.3\%) reduction in Angle errors on the Gaze360, MPIIFaceGaze and RT-Gene datasets, respectively. The parameters are further reduced by 32.9\%, thus proving that Multitask-Gaze can further achieve lightweight while improving accuracy. The performance of Multitask-Gaze is improved for the following reasons:
\begin{itemize}
\item SCA uses an attention mechanism to highlight gaze-related information while enhancing spatial and channel information interactions.
\item UC significantly expands the receptive field of the model and enhances long-distance modeling capabilities, utilizing a wider range of contextual information.
\item GCM further fuses global information and reduces information loss.
\end{itemize}

\begin{table}[]
\caption{Ablation experiment}
\scalebox{0.9}{
\begin{tabular}{llllll}
\hline
Methods        & Params(M) & FLOPs(G) & G\cite{b16} & MPII\cite{b17} & RT\cite{b18} \\ \hline
Multitask-Gaze & 2.828     & 0.243    & 10.33°    & 3.90°        & 6.35°      \\
-MRM           & 2.735     & 0.243    & 10.62°    & 3.88°        & 6.72°      \\
-SCA           & 2.743     & 0.204    & 10.86°    & 3.86°        & 6.55°      \\
-GCM           & 2.816     & 0.250    & 10.73°    & 3.88°        & 6.87°      \\
-ALL           & 2.639     & 0.211    & 11.01°    & 3.84°        & 7.00°      \\ \hline
\multicolumn{4}{l}{- Indicate the removal of the corresponding model.}
\end{tabular}}
\end{table}

\subsection{Ablation experiment}
In order to better understand the effects of different modules on the Angle error, parameter number and FLOPs of the Multitask-Gaze model, the ablation experiment was performed in this section and the results were shown in Table 2. MRM has the largest relative impact on the number of model parameters, while SCA has the largest impact on FLOPs. In addition, the Angle error decreased after the corresponding module was removed from the MPIIFaceGaze dataset, which may have been slightly overfitting due to the small size of the dataset. However, for the other two datasets, there is a large performance degradation after the removal of the corresponding module. When all modules are removed, the performance on the RT-Gene dataset decreases significantly by 10.24\%.

\subsection{Validation of GCM effectiveness}
In order to verify whether GCM can improve the performance of the model, we experimentally replace GCM with global average pooling and max-pooling. The experimental results are shown in Table 3. After replacing GCM with Max-pooling, performance degradation occurred on each dataset. The reason for the performance degradation was that Max-pooling only selected the largest pixel in the feature map for backward propagation, resulting in severe information loss. When the feature map was 5 × 5, the information loss rate was as high as 96\%, and Max-pooling lacked information interaction between channels.
\begin{table}[]
\centering
\caption{Validation of GCM effectiveness}
\begin{tabular}{llll}
\hline
Methods    & G\cite{b16} & MPII\cite{b17} & RT\cite{b18} \\ \hline
GCM         & 10.33°    & 3.90°        & 6.35°      \\
Avg-pooling & 10.73°    & 3.88°        & 6.87°      \\
Max-pooling & 10.73°    & 3.99°        & 6.77°      \\ \hline
\end{tabular}
\end{table}

\subsection{Validation of UC effectiveness}
This section compares the parameter and FLOPs of UC and standard convolution, and visualizes the receptive field using the method proposed by Ding et al. [27]. The data are shown in Table 4 and Fig. 6. Based on Table 4 and Fig. 6, it can be seen that the receptive field size of the 3-layer 5×5 standard convolution and the 3-layer [5×1;1×5] UC are similar, but the UC has 60\% fewer parameters and 39.76\% fewer FLOPs than the standard convolution. However, the receptive field of the [7×1; 1×7] UC is significantly larger and the parameters and computational complexity are 42.5\% and 15.66\% less, respectively, than the standard convolution. The receptive fields of 4-layer 5×5 standard convolution and 3-layer [7×1;1×7] UC are roughly the same, but the parameters and computational complexity of 3-layer [7×1;1×7] UC are significantly reduced by 71.25\% and 19.54\%, respectively. Therefore, larger UC can be used instead of smaller standard convolutions, which not only makes the model lighter, but also significantly expands the receptive field to capture a wider range of contextual information, thereby achieving performance improvement.
\begin{table}[]
\caption{Validation of UC effectiveness}
\scalebox{0.8}{
\begin{tabular}{ccccc}
\hline
Number of layers &           & Standard Convolution & UC             & UC             \\ \cline{3-5} 
                 &           & 5×5                  & {[}5×1; 1×5{]} & {[}7×1; 1×7{]} \\ \hline
3                & Params(M) & 0.04                 & 0.016          & 0.023          \\
                 & FLOPs(G)  & 0.083                & 0.05           & 0.07           \\ \hline
4                & Params(M) & 0.08                 & 0.032          & 0.045          \\
                 & FLOPs(G)  & 0.087                & 0.052          & 0.072          \\ \hline
\end{tabular}}
\end{table}

\begin{figure}[htbp]
\centerline{\includegraphics[width=0.8\linewidth]{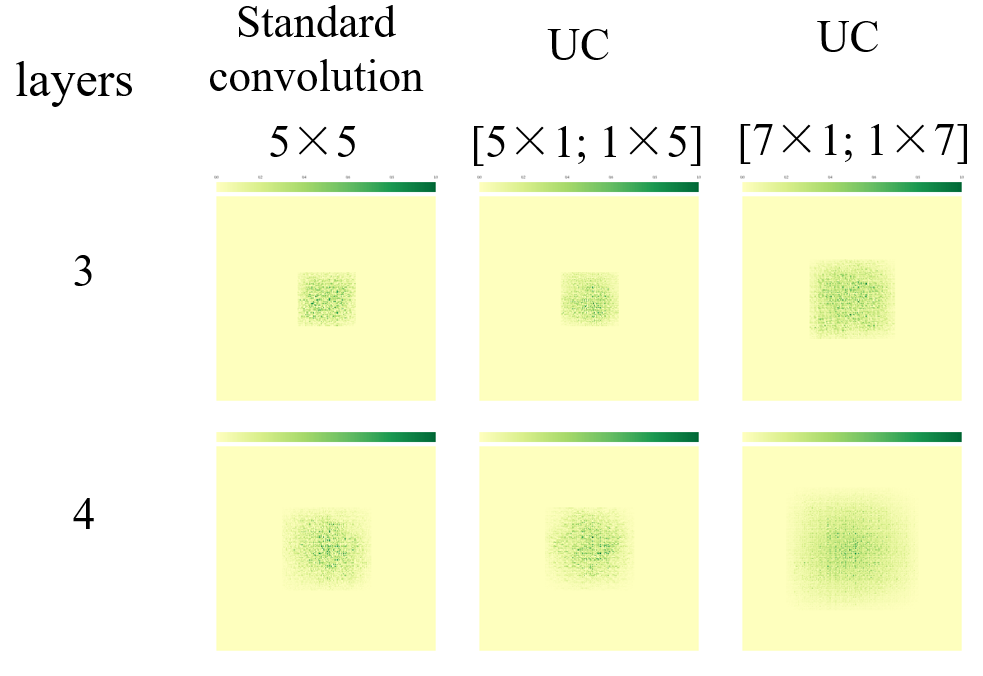}}
\caption{Visualization of receptive field}
\label{receptive field}
\end{figure}

\section{Conclusion}
In this paper, we propose a new lightweight model, Multitask-Gaze, and verify the model performance on a widely used dataset. The Multitask-Gaze is not only more lightweight, but also significantly improves performance. The advantages of the SCA, UC, GCM and MRM modules presented in this paper are strongly demonstrated, where SCA and GCM can achieve plug-and-play effects and provide new ideas for model lightweighting.

\vspace{12pt}
\color{red}
%IEEE conference templates contain guidance text for composing and formatting %conference papers. Please ensure that all template text is removed from your %conference paper prior to submission to the conference. Failure to remove the %template text from your paper may result in your paper not being published.

\end{document}